\documentclass[final]{anthology-ch} % for the final version
\usepackage{booktabs}
\usepackage{enumitem}
\usepackage{listings}

\usepackage{graphicx}
\usepackage{tcolorbox}
\usepackage[capitalise,noabbrev]{cleveref}
\usepackage{subcaption}

\usepackage{siunitx}
\usepackage{pifont}
\usepackage{booktabs}
\usepackage{tabularx}
\usepackage{array}
\newcolumntype{Y}{>{\raggedright\arraybackslash}X}

\tcbset{
  modernprompt/.style={
    colback=gray!5,
    colframe=gray!20,
    coltitle=black!90,       % light grey title text
    fonttitle=\bfseries,    % keep title bold
    boxrule=0.5pt,
    arc=3pt,
    outer arc=3pt,
    sharp corners=south,
    top=2pt, bottom=2pt, left=6pt, right=6pt,
    boxsep=4pt,
    attach title to upper=false,  % don’t reserve a title slot
    title={},                     % clear any title text
    coltitle=.,                   % (optional) no title colour
    fonttitle=\relax 
  }
}

\crefname{tcolorbox}{Prompt}{Prompts}
\Crefname{tcolorbox}{Prompt}{Prompts}

\usepackage[abbreviations,automake,nonumberlist]{glossaries-extra}
\newabbreviation{om}{OM}{Ontology Matching}
\newabbreviation{dh}{DH}{Digital Humanities}
\newabbreviation{skos}{SKOS}{Simple Knowledge Organization System}
\newabbreviation{mt}{MT}{Machine Translation}
\newabbreviation[shortplural=LLMs, longplural=Large Language Models]{llm}{LLM}{Large Language Model}
\newabbreviation{oaei}{OAEI}{Ontology Alignment Evaluation Initiative}
\newabbreviation[shortplural=RQs, longplural=research questions]{rq}{RQ}{research question}

\makeglossaries[abbreviations]

\title{Mind the Language Gap in Digital Humanities: LLM-Aided Translation of SKOS Thesauri}

\author[1]{Felix Kraus}

\author[1]{Nicolas Blumenröhr}

\author[1]{Danah Tonne}

\author[1]{Achim Streit}

\affiliation{1}{Scientific Computing Center, Karlsruhe Institute of Technology, Karlsruhe, Germany}


\pubyear{2025}
\pubvolume{1}
\pagestart{1}
\pageend{1}
\conferencename{Proceedings of Conference XXX}
\conferenceeditors{Editor1 Editor2}
\doi{00000/00000}  

\begin{document}

\maketitle

\begin{abstract}
% the abstract should be a one-paragraph summary of the outline and main contributions of the paper.
We introduce WOKIE, an open-source, modular, and ready-to-use pipeline for the automated translation of SKOS thesauri. This work addresses a critical need in the Digital Humanities (DH), where language diversity can limit access, reuse, and semantic interoperability of knowledge resources. WOKIE combines external translation services with targeted refinement using Large Language Models (LLMs), balancing translation quality, scalability, and cost. Designed to run on everyday hardware and be easily extended, the application requires no prior expertise in machine translation or LLMs. We evaluate WOKIE across several DH thesauri in 15 languages with different parameters, translation services and LLMs, systematically analysing translation quality, performance, and ontology matching improvements. Our results show that WOKIE is suitable to enhance the accessibility, reuse, and cross-lingual interoperability of thesauri by hurdle-free automated translation and improved ontology matching performance, supporting more inclusive and multilingual research infrastructures.
\end{abstract}

\section{Introduction and Motivation}\label{sec:introduction}
The organization and structuring of knowledge relies on controlled vocabularies, thesauri\footnote{A thesaurus arranges terms based on synonymy, hierarchical relationships and other properties. In contrast to ontologies, which offer comprehensive structures for semantic reasoning, thesauri primarily standardize terms.}, and ontologies. These resources support the use of software for querying linked data, annotating datasets, and even forming core research objectives \cite{haslhoferKnowledgeGraphsLibraries2018}. Within the \gls{dh}, research communication and (metadata) publication frequently occur in English. This creates barriers for non-native English speakers and can limit or exclude them from participation and access to knowledge \cite{violaEditorialDataWorkflows2024}. As a result, the culture, language, and history of these communities are under- or misrepresented, leading to a loss of cultural diversity and richness \cite{violaEditorialDataWorkflows2024}. An important element to lower the barrier is to include additional languages in thesauri, particularly those relevant to the regions or communities connected to the research objects. Multilingual thesauri, often used for metadata, enable \textit{collective benefit}, especially for those communities directly connected to the research, aligning with the \textit{C} of the CARE principles\footnote{\url{https://www.gida-global.org/care}}.

As efforts by different communities grow to create multilingual thesauri, content overlaps are inevitable \cite{euzenatOntologyMatching2013}. Additionally, different platforms often require slightly different terminologies, further contributing to knowledge fragmentation \cite{morvilloIntegratingMultipleKnowledge2024}. \Gls{om} addresses this heterogeneity by aligning equivalent terms or rather concepts\footnote{Although the used data model (SKOS) only defines concept as "an idea or notion; a unit of thought" \cite{milesSKOSSimpleKnowledge2009}, we use "term" instead within this work to avoid ambiguity.}. However, multilingual thesauri present specific challenges for existing \gls{om} systems, particularly in the \gls{dh}, where non-English content, historical languages, and various scripts are prevalent. Due to the predominance of English across most domains, current \gls{om} systems perform poorly on non-English or multilingual thesauri using the widespread  \gls{skos} model. The addition of English labels before matching is promising, which enables better performance of \gls{om} systems. 

Achieving the translation step manually, whether for enabling multilingual access or improving \gls{om}, is not scalable. The use of classical statistical or neural  \gls{mt} methods for automated translation demands extensive bilingual corpora for supervised learning or monolingual data for unsupervised learning. Both approaches face significant limitations in \gls{dh} contexts, where domain-specific bilingual corpora and training data are scarce \cite{suissaTextAnalysisUsing2022}, and unsupervised methods yield poor results in cases of domain mismatch and low-resource languages \cite{kimWhenWhyUnsupervised2020}. 

Given these limitations, \glspl{llm} have emerged as a promising method for translation \cite{jiaoChatGPTGoodTranslator2023}. Generic models do not require training by the individual researcher, but have the downside of longer processing times, possible hallucinations, and non-deterministic results. To counterfeit these issues while ensuring high-quality translations, we propose combining multiple external translation services with LLM-based refinement. Currently, no dedicated pipeline exists that integrates these two approaches specifically for translating \gls{skos} thesauri. 

To fill this gap, we introduce WOKIE (\textbf{W}ell-translated \textbf{O}ptions for \textbf{K}nowledge Management in \textbf{I}nternational \textbf{E}nvironments) which balances throughput, cost, and quality. WOKIE selectively applies \gls{llm} refinement when translations of multiple external services disagree.  It uses configurable thresholds to only use \glspl{llm} when necessary,  also reducing potential hallucinations by comparing \gls{llm} outputs directly to existing translation candidates. WOKIE is written in python, runs on everyday hardware, and can use solely free translation services. Translations generated by WOKIE are immediately integrated in the thesaurus, making the pipeline accessible to \gls{dh} researchers without specialized infrastructure or extensive technical expertise. 

The design of WOKIE is modular and easy-to-use: it applies a user-selected combination of eight primary translation services like Google translator, Lingvanex and ModernMT. Optionally, additional services can be implemented by using a common interface. The utilized \gls{llm} is chosen by the user out of 27 implemented models. Moreover, new models can be added easily, for example to use free or low-cost models provided by research institutions. This enables to tailor WOKIE to specific use cases. 
% While the evaluation focuses on specific \gls{dh} domains, WOKIE itself relies on general-purpose translators and should be suitable for domains outside the \gls{dh}.

In this work, we present the following main contributions, including an evaluation structured around five \glspl{rq}:
\begin{itemize}
    \item \textbf{WOKIE}, an open-source\footnote{\url{https://github.com/FelixFrizzy/WOKIE}}, modular and ready-to-use pipeline for automatic translation of \gls{skos} thesauri, designed to run on everyday hardware and to be easily extendable.
    \item \textbf{RQ 1:} Which external translation services are most suitable as primary translators for the pipeline?
    \item \textbf{RQ 2:} How does LLM-based refinement impact the translation quality of \gls{skos} thesauri?
    \item \textbf{RQ 3:} Which configuration parameters yield the best translation quality?
    \item \textbf{RQ 4:} Which \glspl{llm} are giving best results when used in the translation pipeline?
    \item \textbf{RQ 5:} What is the impact of pre-translation on ontology matching results?
\end{itemize}

\section{Related Work} \label{related-work}
While no translation system specifically targets \gls{skos} thesauri, several methods exist for translating ontologies and knowledge bases. Besides manual translation, different types of \gls{mt}, including neural approaches, have been developed. Since they might be adoptable to the translation of \gls{dh} \gls{skos} thesauri, we review them below. 

Early work on ontology label translation combined web services with contextual information from a database of multilingual ontologies \cite{espinozaLabelTranslatorToolAutomatically2008}. It uses a semi-automatic approach presenting the user ambiguous cases, which is not scalable. More recently, a simpler approach without contextual information was implemented as a plugin for the ontology editor Protégé\footnote{\url{https://protege.stanford.edu/}} \cite{florrenceMLGrafVizMultilingualOntology2021}, where disambiguation is not possible.

Another approach for translation is rule-based \gls{mt}, relying on lexicons and rules defined by linguists \cite{thurmairComparingRulebasedStatistical2004}. In contrast, statistical \gls{mt} and example-based \gls{mt} methods learn from existing translations collected in bilingual corpora that include the domain of the translation task \cite{moussallemMachineTranslationUsing2018}. Statistical \gls{mt} has been further enhanced by adding contextual information from lexicons \cite{mauserExtendingStatisticalMachine2009} or parallel corpora \cite{arcanOntologyLabelTranslation2013}, which showed slight improvements over standard statistical \gls{mt}. These approaches require well-prepared datasets built by linguists or \gls{mt} experts. In our case, such resources are scarce, and we aim for a method that can be applied by \gls{dh} domain experts without extensive data preparation or detailed knowledge of machine translation.

Adaptive neural networks have also been proposed for knowledge base translation \cite{fengEnglishChineseKnowledgeBase2016}. They represented subject and object triples as word embeddings and mapped them into a shared vector space using an embedding learning algorithm. Translation candidates obtained by an external service are ranked using these embeddings, making results highly dependent on external service accuracy. 
Furthermore, Neural \gls{mt} requires pre-aligned training data for accurate results. Combining text- and triple-based models has shown better performance compared to the latter alone \cite{moussallemTHOTHNeuralTranslation2019}. For the medical and financial domain, it has been shown that Neural \gls{mt} models can be better fine-tuned for specific domains, leading to improved results compared to statistical \gls{mt} \cite{arcanTranslatingTerminologicalExpressions2019}.

To apply Neural \gls{mt} to the \gls{dh} domain where training data are scarce, architectural changes or adaptions are needed to achieve the same level of accuracy than in domains with more available training data \cite{suissaTextAnalysisUsing2022}. For instance,  intermediate fine-tuning steps have improved Neural \gls{mt} results for French-Dutch translation in  Fine Arts \cite{banarTransferLearningDigital2020}. Unsupervised Neural \gls{mt} avoids the need for parallel corpora, but the performance is affected whenever source and target monolingual data show linguistic differences and domain mismatch or low-resource languages are involved \cite{kimWhenWhyUnsupervised2020}.

More recently, \glspl{llm} such as GPT-3 have been used for zero-shot translation. Due to their training on massive multilingual datasets \cite{brownLanguageModelsAre2020}, these models require no additional fine-tuning and are therefore accessible even for non experts in \gls{mt}. GPT-4.5 for instance has demonstrated competitive results, outperforming some specialized neural \gls{mt} systems, even leading German-English and English-German translation tasks \cite{manakhimovaLinguisticallyMotivatedEvaluation2023}. Similarly, ChatGPT achieves translation quality comparable to commercial services, even for language pairs with substantial typological and syntactic divergence, such as Chinese and Romanian \cite{jiaoChatGPTGoodTranslator2023}. Their accessibility, broad language coverage and competitive results make them particularly relevant for our scenario. Therefore, they are examined in more detail in this work. 

There exist also fine-tuned models for text-to-text translation, such as mT5 \cite{xueMT5MassivelyMultilingual2021} and mBART \cite{liuMultilingualDenoisingPretraining2020}. However, they typically translate entire text inputs at once, making them unsuitable for translating individual terms enriched with context for disambiguation.

\section{Translation Pipeline}
\subsection{Design Choices}
WOKIE is a lightweight, modular pipeline designed to run efficiently on everyday hardware. The terms are translated individually, first with a user-selected set of external translation services, the primary services, which are generally fast and cost-efficient, or free. When necessary, an interchangeable \gls{llm} is employed for further refinement. It provides context-sensitive translation, but comes with higher latency, token-based costs, and usage limits. We utilize zero-shot prompts for the \gls{llm} to keep WOKIE domain-agnostic and avoid complex fine-tuning, especially given the challenges in \gls{dh}. 

All implemented translators follow a common base class, which enables the easy integration of new translation services or \glspl{llm}. The supported languages depend solely on the chosen services, e.g. Google Translate supports 244 different languages, including Latin. We focus on the translation of \texttt{skos:prefLabel}, but other properties can be selected easily. The confidence threshold triggering \gls{llm} refinement is configurable: a higher threshold results in more frequent \gls{llm} usage, which increases execution time and potentially costs. The resulting translations are serialized back into an enriched thesaurus, enabling seamless use as a preprocessing step for any thesaurus creation, management, or alignment software.

\subsection{Pipeline Overview}
\begin{figure}
    \centering
    \includegraphics[width=1\linewidth]{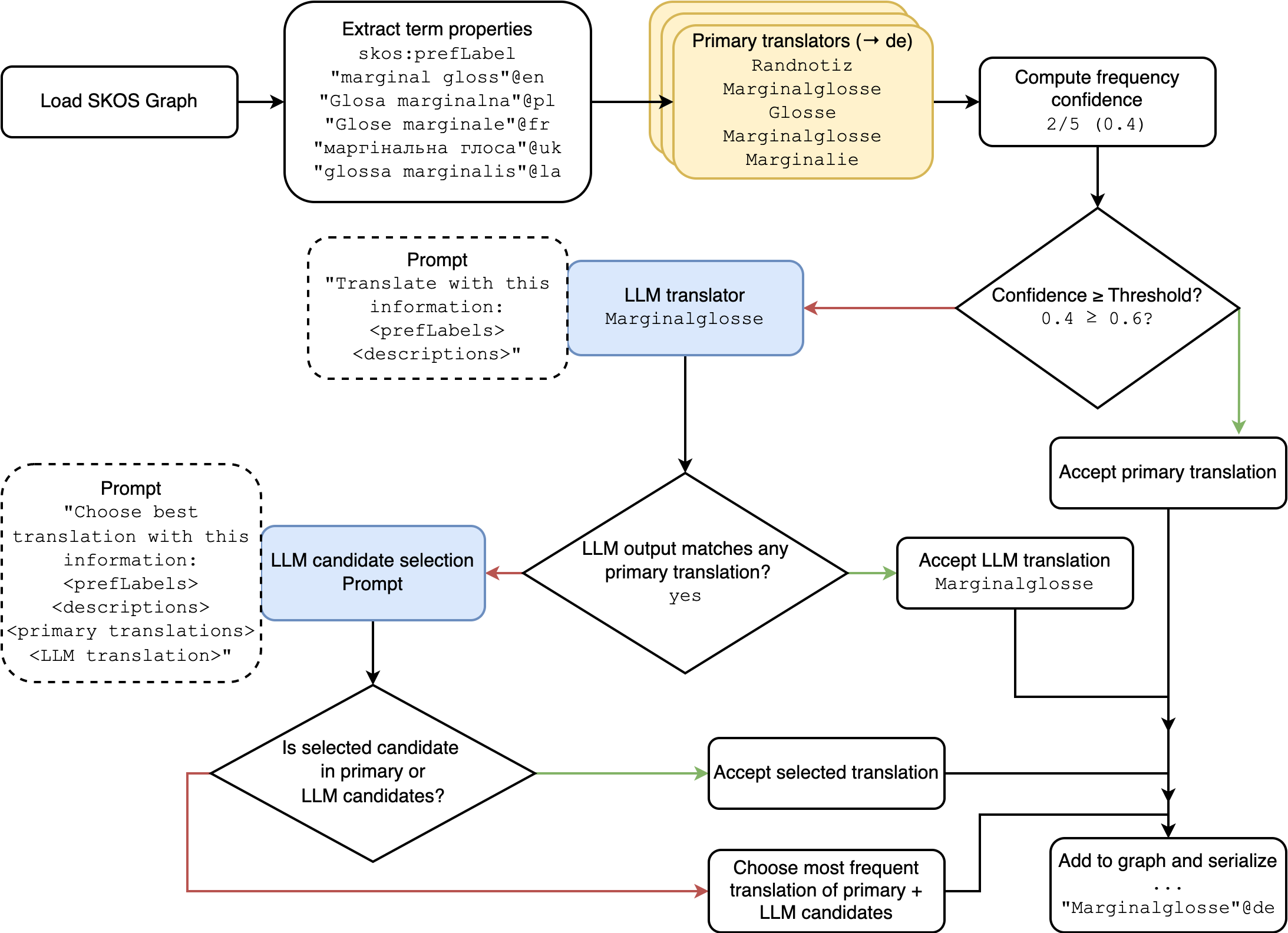}
    \caption{Simplified flowchart diagram of WOKIE's translation steps using "marginal gloss" as an example. The involvement of external translation services is marked in yellow, the use of \glspl{llm} blue. Positive decisions are on the right-hand side of the diamond (green arrow), negative decisions on the left (red arrow). }
    \label{fig:workflow}
\end{figure}

WOKIE processes a \gls{skos} thesaurus using three main components: primary translators, confidence calculators and LLM-based refinement. A high-level overview of this pipeline is depicted in \cref{fig:workflow}. For each term and selected property (default is \texttt{skos:prefLabel}), WOKIE first collects primary translation candidates from the primary translation services. It then calculates a simple frequency-based confidence score. If the score meets or exceeds a user-defined threshold, the most frequent candidate is accepted directly. Otherwise, the \gls{llm} is used to get a context-sensitive secondary translation. If this matches a primary candidate, it is chosen as the final translation. If not, the \gls{llm} selects the best option of all primary and secondary candidates. The final translation is added back into the \gls{skos} graph as a new literal in the target language. 

\subsubsection{Primary Translations} \label{primary-translations}
Users can select one or more primary translation services and prioritize their order. Each language label of a term is translated independently. Multiple services can be used to meet the minimum number of translations specified by the user. 
For monolingual thesauri, the maximum number of obtainable candidates equals the number of services that support the source-target language pair. The pipeline ensures translation of all labels, even if the minimum required translations are fewer. This ensures that as much information as possible is used for the translations.

\subsubsection{Frequency-Based Confidence}
A simple frequency method selects the most common translation based on exact string matches. Confidence is calculated as the number of occurrences of this translation divided by the total number of candidates. If the confidence is greater than or equal to the threshold, the translation is accepted immediately, bypassing the \gls{llm} refinement to save resources. Ambiguous cases, however, trigger further refinement. 

\subsubsection{LLM-Based Translation}
Following the conclusions of \S\ref{related-work}, we make use of \glspl{llm} to support disambiguation. When primary confidence is too low, a prompt is constructed that includes the term description or definition, the general vocabulary description or a user selected context, depending on availability, for example:

\begin{tcolorbox}[modernprompt, title=]

\emph{Instructions}: You are a machine translation system that translates a term from any language to English.  

To determine the correct context, use the provided additional details. Return only the translated term and nothing else.

\emph{Input}: Term to translate: marginal gloss  

Description of the term that should be translated: 

A marginal gloss is a brief note in the margin explaining text.
\end{tcolorbox}

While some models are verbose, the translation is still extracted if possible. To make it resilient to hallucinated or inadequate translations, a final translation is only accepted if it matches any of the primary candidates.

\subsubsection{LLM-Based Candidate Selection}
If the LLM's translation is not in the primary translations, a prompt is used to choose the most accurate translation among all candidates. Parts about output format are spared out in the example:

\begin{tcolorbox}[modernprompt, title=]

\emph{Instructions}: You are a professional translation review system that assesses the quality of translations of a single term given in different source languages. The translations are already given by a translation system. Give me the best fitting translation out of the given list (...)

%and a confidence how sure you are that the translation is accurate on a scale from 0 to 1. If no possible translation seems to be fitting, return None as best fitting translation and a confidence of 0.

Criteria for high accuracy are:
\begin{itemize}
    \item The best fitting translation is already found in the already given possible translations
    \item In the current context, there is no possible translation that has a different meaning.
\end{itemize}
Only give me the best fitting translation (...)

% and the confidence in this format:
    % best fitting translation; confidence
% Return only the best fitting translation and confidence.

\emph{Input}: Choose the best fitting translation to German. 
(\emph{first part identical to translation prompt})

The possible translations to German coming from translation systems are: 

Randnotiz, Marginalglosse, Glosse, Marginalie

(...)
% Return the best fitting translation and the confidence in this format:

% best fitting translation; confidence
\end{tcolorbox}
The following steps are similar to the \gls{llm} translation step. If parsing fails or the selection is invalid, a simple frequency-based calculation is used as a fallback option.

\section{Evaluation Methodology}
The primary goal of the evaluation is to compare various external translation services and \glspl{llm} and assess their suitability for translating \gls{dh} thesauri within our pipeline. This goal is structured around the research questions posed at the end of \S\ref{sec:introduction}.

\subsection{Data Preparation and Translator Selection}
As a first step, we identified and implemented 28 external translation services that provide an API, using wrapper libraries whenever possible. Due to issues such as low request limits, slow and unreliable responses, and other errors, eight services remained suitable for the evaluation. For selecting multilingual \glspl{llm}, we followed a similar process and also included models of a benchmark leaderboard \cite{zhou2024multilingual} if possible. A comprehensive overview can be found in our GitHub repository\footnote{\url{https://github.com/FelixFrizzy/WOKIE/blob/main/supported-services.md}} or in \cref{appdx:tables}. 

For the accuracy comparison, we selected different multilingual \gls{dh} thesauri, partly based on an existing multilingual \gls{dh} ontology matching benchmark \cite{krausGoldStandardBenchmark2024}, see \cref{tab:thesauri-resources}. We selected them because they are suitable for evaluating the effect of translation on ontology matching, and their size allows for repeated runs with different configurations and manual review. These thesauri use fifteen different languages across four scripts: Latin, Cyrillic, Greek, and Arabic. For each test case, we removed all properties in one language from a thesaurus and used WOKIE to translate them back. Then we compared the back-translated terms to the originals. When choosing which languages to remove, we excluded the language in which the thesaurus was originally developed. We determined this by examining early versions, related projects, or contacting developers. This avoided back-translating  \gls{mt}-generated label, preventing bias. 

\begin{table}[ht!]
\centering
\small
\makebox[\textwidth]{ % This will center the minipage
\begin{minipage}{\textwidth}
\centering
\scriptsize
\begin{tabular}{llll}
\hline
Resource & Version / Date & \#terms\footnotemark{} & language (ISO 639)\\
\hline
CodiKOS & - / - & $\sim$170 & de, en \\
\hline
DEFC Thesaurus & - & $\sim$800 &  de, en, la \\
\hline
DYAS & 3.1 / 2020-10-21 & $\sim$30 & de, el, en, fr, la \\
\hline
iDAI.world Thesaurus & 1.2 / 2022-02-10 & $\sim$270 & de, en, es, fr, it \\
\hline
Iron-Age-Danube thesaurus & 1 / 2018-11-07 & $\sim$290 & de, en, hr, hu, sl \\
\hline
OeAI Thesaurus - Cultural Time Periods & 1.0.0 / 2022-11-23 & $\sim$400 & de, en \\
\hline
TaDiRAH & 2.0.1 / 2021-07-22 & $\sim$170 & de, en, es, fr, it, pt, sr \\
\hline
UNESCO & - / 2024-06-03 & $\sim$490 & ar, en, fr, es, ru \\
\hline
\end{tabular}
\end{minipage}
} % End of makebox
\caption{Thesauri used for the dataset.}
\label{tab:thesauri-resources}
\end{table}
\footnotetext{Since we only used multilingual concepts, this number might be smaller than the unaltered thesaurus.}

\subsection{Accuracy Measures}
We used four different similarity measures to compare the original term with the back-translation, each normalized between 0 and 1, with the latter meaning identical strings. These measures provide relative comparisons rather than absolute similarity which is sufficient for our comparative analysis.
\paragraph{Exact Match} A boolean test of string equality ignoring the case was used to find exact matches. This represents a conservative estimate of translation accuracy. 
\paragraph{Levenshtein Similarity} The Levenshtein or edit distance \cite{levenshteinBinaryCodesCapable1966} counts the minimum number of character operations (removal, insertion, substitution) required to transform one string into another. 
\paragraph{Jaro-Winkler Similarity} The Jaro similarity \cite{jaroAdvancesRecordLinkageMethodology1989} considers the number of common characters and transpositions between two strings, useful for measuring lexical similarity. The Winkler modification \cite{winklerStringComparatorMetrics1990} improves the quality \cite{christenComparisonPersonalName2006} by increasing the Jaro similarity when initial characters (up to four) match. 
\paragraph{Cosine Similarity} We chose the pre-trained BPEmb model \cite{heinzerlingBPEmbTokenizationfreePretrained2018}, representing each word as an average of its subword vectors. % vocabulary size hab ich rausgelassen, hätte ich begründen können mit heinzerlingSequenceTaggingContextual2019, ist aber auch nicht relevant für das paper
Cosine similarity measures the semantic similarity based on the angle between the vectors. The smaller this angle, the higher the semantic similarity. BPEmb is best suited for our case because, for one part, using subword vectors prevents out-of-vocabulary errors that occur frequently on \gls{dh} technical terms using word-based vectors. For the other, it is available in 275 languages, including Latin. Additionally, it  has an over 540-times lower memory footprint than the comparable multilingual model fastText \cite{heinzerlingBPEmbTokenizationfreePretrained2018, bojanowskiEnrichingWordVectors2017}. This makes it ideal for running on everyday hardware, allowing users to perform tailored comparisons. LLM-based embeddings such as Gemini Embedding \cite{leeGeminiEmbeddingGeneralizable2025} were excluded to maintain independence from the evaluated \glspl{llm}.

\subsection{Test Procedure}
For each evaluation task, we selected thesauri samples across different languages and systematically varied one of the following parameters, keeping others constant: 
\begin{itemize}
    \item primary translation service,
    \item LLM,
    \item prompt composition,
    \item \gls{llm} temperature,
    \item confidence threshold, and
    \item minimum number of translations.
\end{itemize}

To ensure stability and reproducibility, some pipeline configurations were executed multiple times. We observed only negligible variations that have no impact on the results or their interpretation.

\section{Evaluation Results}
All scripts and thesauri used for the evaluation, their translations under various settings, computed distance measures, and related metadata are openly available as Zenodo record\footnote{\url{https://doi.org/10.5281/zenodo.15494760}}. An example of translated labels can be found in \cref{appdx:tables}. The basic requirements for the evaluation of translation services and \glspl{llm} are a stable API, adequate request limits and reasonable output. 

\subsection{Comparison of Primary Translation Services} \label{primary-translation-services}
\begin{figure}
  \centering
    \includegraphics[height=6.2cm]{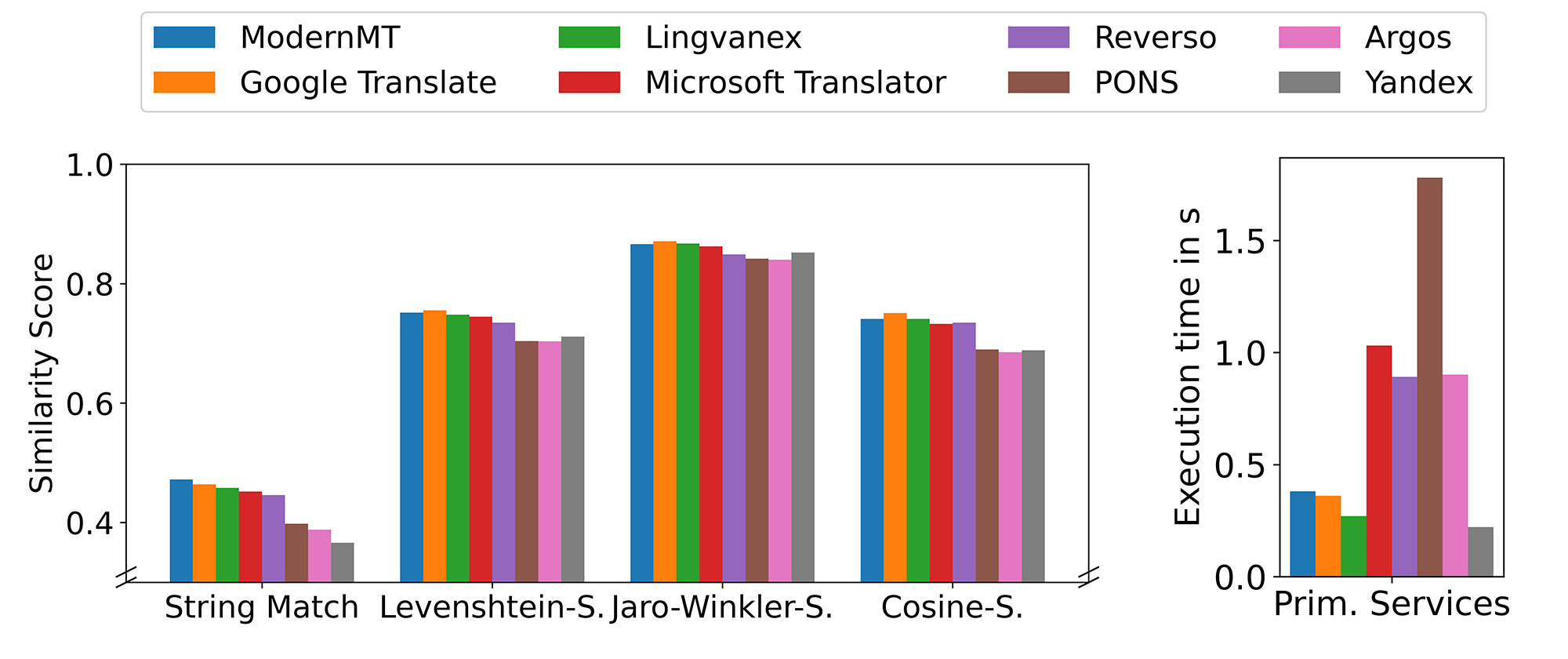}
    % \caption{Primary services evaluation}
  \caption{Comparison of the performance (left) and execution time (right) of different primary services, averaged over test cases.}
  \label{fig:primary-comparison}
\end{figure}

To address the first research question, we used 14 test cases consisting of thesauri that included only languages supported by all primary translation services. \cref{fig:primary-comparison} shows the macro average over all test cases (left) and micro averaged execution time per translation (right). Looking at string similarity, PONS, Argos, and Yandex consistently underperformed compared to other services. Similar is true for Levenshtein- and cosine similarity, although Yandex achieved better results for Jaro-Winkler similarity. For Latin translations, Google Translate performed best, with Lingvanex reaching between 75 and 93\%, and ModernMT between 25 and 76\% of Google's performance. 

Comparisons across languages are challenging because the domain influences the results. However, looking at a single thesaurus only, comparisons are possible. For multilingual TaDiRAH, German and Serbian were the most difficult to translate accurately. For Latin within the DEFC Thesaurus, the string matches dropped to 2.38\%, although cosine similarity was still at 0.25 which shows that translating Latin is possible, although it is likely that there is a bias introduced because Latin was most probably not the language in which it was developed. 

Regarding execution times, the slowest, PONS, was eight times slower than Yandex with \SI{0.22}{\second} per translation. The execution times were generally consistent across languages for single services. 

Due to occasional request limits encountered with ModernMT and Microsoft Translator, and to manage the usage of Google's paid service efficiently, we recommend the following prioritization for optimal translation outcomes: Lingvanex, Google Translate, ModernMT, Microsoft Translator, Yandex, Argos, Reverso, PONS.

\subsection{Prompt Engineering}
During our prompt experiments, several \glspl{llm} repeatedly failed to provide valid translations, instead generating irrelevant content such as repeating our question,  definitions, or programming code. These models, of which most were free or older models like Llama or Mistral / Mixtral,  were disregarded for further experiments. For models offering separate instruction and input prompts, we discovered that repeating instructions in the input leads to a more streamlined output. 

We also evaluated three prompt strategies: individual, batch, and hierarchy. The individual approach translates all multilingual labels of one term individually, resulting in multiple translation candidates per term. The batch approach combines the information coming from all term labels into one prompt, producing a single translation candidate per term. The hierarchy approach expands the individual method by additionally including all broader terms up to the root term in the prompt.

None of the prompt strategies outperformed the others. The hierarchical method increased token usage without benefits because none of the tested representations of hierarchy relations could be correctly processed by the LLMs and quality did therefore not improve. The batch prompts provided only one translation per term, leading to less resilience to errors. We therefore chose the individual prompt approach for its simplicity and robustness.

\subsection{Impact of Refinement}
To assess the impact of \gls{llm} refinement (RQ2), we translated six thesauri using different pipeline configurations. Preliminary tests indicated that Gemini 2.0 Flash consistently ranked among the best-performing \glspl{llm}. We selected it as a representative model to ensure a fair comparison across the following pipeline configurations:
\begin{itemize}
    \item single primary translator only,
    \item all recommended primary translators (see \S\ref{primary-translation-services}),
    \item single \gls{llm} only, no primary translators, and
    \item combination of all primary translators with \gls{llm} refinement.
\end{itemize}

\begin{figure}
    \centering
    \includegraphics[width=1\linewidth]{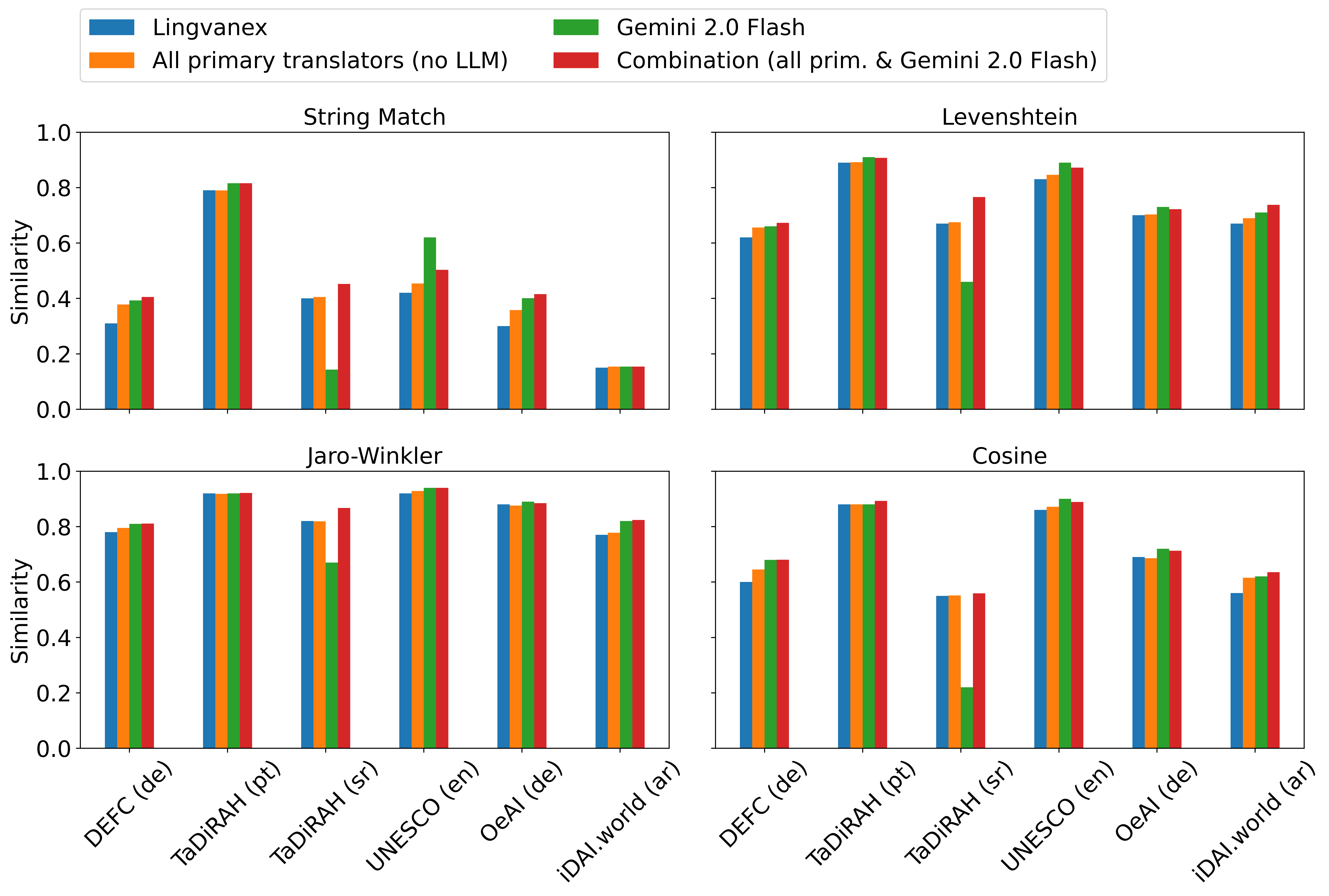}
    \caption{Similarity measures for the four test setups and multiple test cases. The removed and back-translated language is indicated in parentheses.}
    \label{fig:LLM}
\end{figure}

Results in \cref{fig:LLM} demonstrate that relying solely on one of the best-performing primary translators, Lingvanex, yields lower accuracy than other approaches. For configurations using all primary translators, we set the minimum number of required translations to five. This means that the pipeline stops querying once five translations are collected (or all labels are translated at least once), which does not always require all services to be called. In monolingual thesauri such as DEFC\_de, this often leads to querying five different services, since only one source label is available per term. In contrast, multilingual thesauri like TaDiRAH\_pt already contain labels in different languages. In such cases, a single service is sufficient to meet the minimum, as it can translate all labels across languages. As a result, the difference in translation quality between using one and all primary services is much smaller for these multilingual cases.

Most importantly, the results strongly suggest that either solely LLM-based translation or combining all primary translators with \gls{llm} refinement provides the best overall results. Surprisingly, Gemini 2.0 Flash underperforms when only used for translation to Serbian, but used just for refinement of difficult cases works still fairly well. Given that \glspl{llm} typically have higher latency, cost, and unclear training language coverage, the optimal configuration uses external translation services with targeted \gls{llm} refinement for challenging cases.

Across test cases, the four similarity measures tend to align in relative terms: when one measure indicates low or high similarity, the others usually follow the same trend. This consistency, though not exact in absolute values, supports the qualitative reliability of the findings.

\subsection{Parameter Tuning}
To determine the optimal parameter set (RQ3), we conducted a series of experiments. For each, we used a set of base settings: a threshold of 0.6, a minimum of five primary translations, a temperature of 0 and using the individual prompt strategy. We selected six \glspl{llm} from six different providers that had shown promising results in preliminary tests, based on translation quality, rate limits, and token pricing. We also reduced the number of test cases to three to allow a manageable amount of different runs. 

\paragraph{Temperature}
The temperature setting influences the randomness  and creativity of an LLM's output. Therefore, we tested the impact of temperature values 0 (close to deterministic), 0.5 and 1 (more variable) on the translation results. Across most models, we observed no notable performance differences. This is likely due to the short length of the output, which limits the model’s opportunity for variation. Our findings align with the literature, where no significant impact on problem-solving tasks within this temperature range was reported \cite{renzeEffectSamplingTemperature2024}. 

\begin{figure}
    \centering
    \includegraphics[width=1\linewidth]{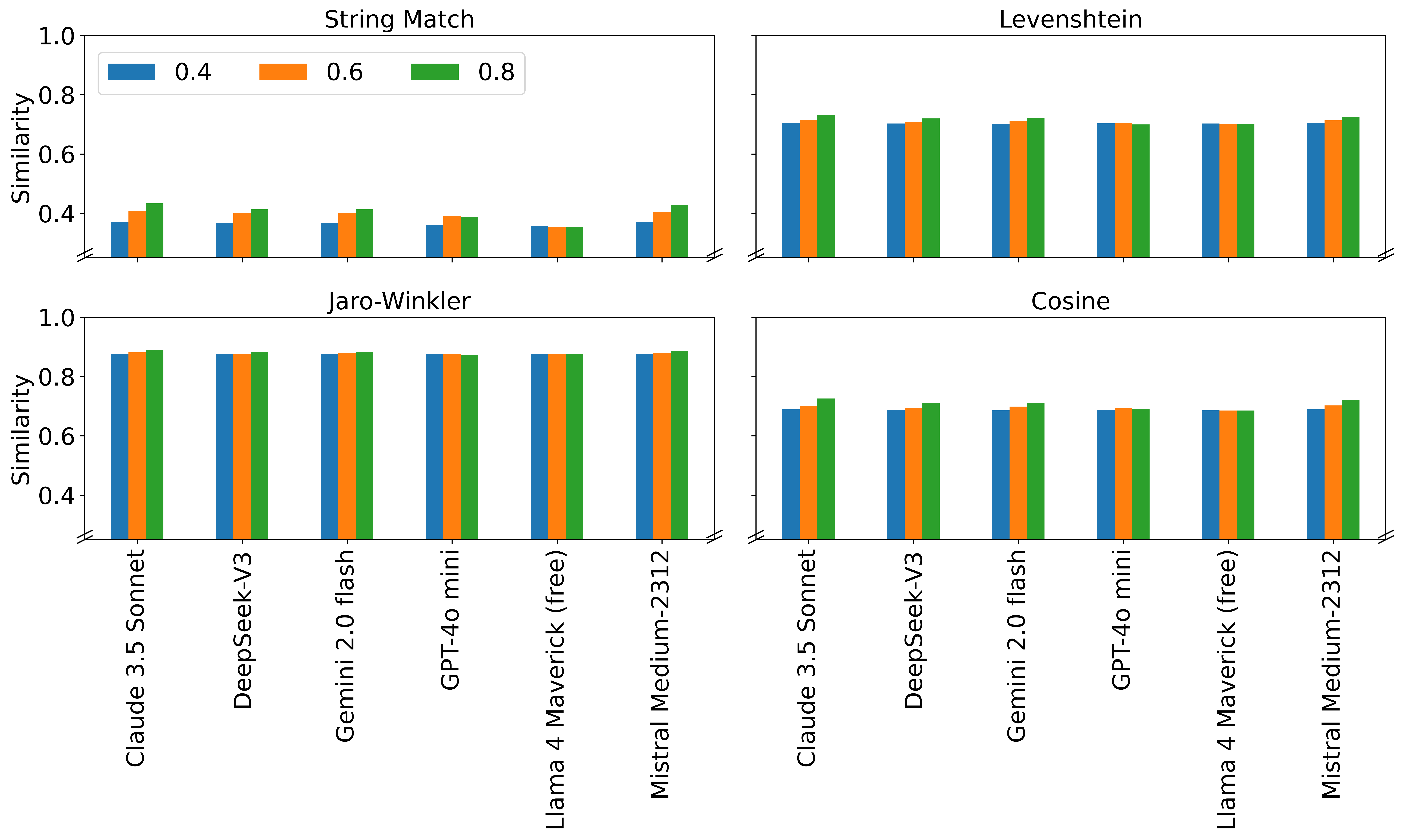}
    \caption{Similarity measures for different threshold values using OeAI Thesaurus back-translated to German.}
    \label{fig:threshold}
\end{figure}

\paragraph{Threshold}
The confidence threshold is closely tied to the number of primary translations, which is kept at five for these experiments. We chose to set the threshold to 0.4, 0.6 and 0.8, which corresponds to accepting a translation directly if at least two, three, or four out of five primary translation candidates agree. As shown in \cref{fig:threshold}, increasing the threshold slightly improved translation quality for some models. A similar conclusion can be drawn for tests with other thesauri.

The higher the threshold, the more often the \gls{llm} is called. The total running time, which is proportional to the number of \gls{llm} calls, rises only slightly when raising the threshold from 0.4 to 0.6. However, moving to 0.8 significantly raised execution time by roughly 50--100\%, depending on the model. Given the minimal quality improvement beyond 0.6, this higher cost is not justified. We therefore selected 0.6 as the optimal threshold.

\paragraph{Minimum Number of Primary Translations}
As mentioned in \S\ref{primary-translations}, users can set the minimum number of primary translations, although the pipeline always ensures that at least all labels of a term are translated once. Because this parameter is linked to the confidence threshold, isolating its effect was not possible. We therefore only performed selected tests with values of three, five, and eight. These sometimes showed either a slight decrease in accuracy for values of three or eight, depending on the test case and language model. Therefore, we selected the minimum number of primary translations to five, which consistently showed good results. We conclude that at least three identical translation candidates are recommended for direct acceptance as final translation. This is the sweet spot between accuracy and resources, so the product of threshold and minimum translations should be no less than three.

\subsection{LLM Comparison}
\begin{figure}
    \centering
    \includegraphics[width=1.0\linewidth]{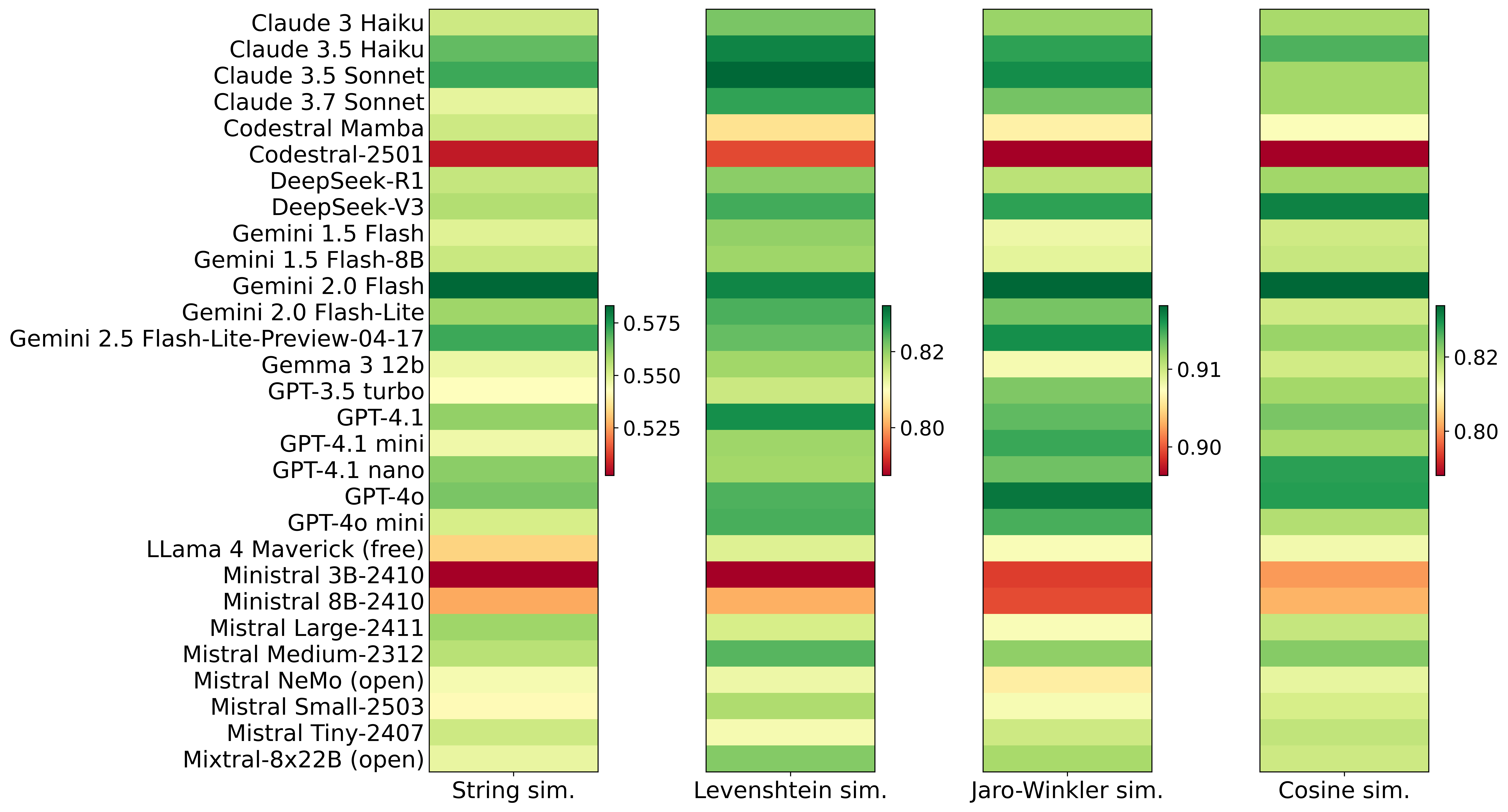}
    \caption{Similarity measures for different \glspl{llm} averaged over test cases. Green indicates the best values, transitioning over yellow to red for lower scores. }
    \label{fig:llm-comparison}
\end{figure}

Having shown the general benefit of LLM-based refinement,  the fourth research question about the best fitting models becomes apparent. We used the optimal parameters and translation services identified earlier for this evaluation. \cref{fig:llm-comparison} presents the averaged results across several test cases. Additional details on pricing and availability can be found in the GitHub repository\footnote{\url{https://github.com/FelixFrizzy/WOKIE/blob/main/supported-services.md}}. 

The similarity scores indicate a group of well-performing models: Gemini 2.0 Flash, DeepSeek-V3, GPT-4o, Claude 3.5 Haiku, Gemini 2.5 Flash-Lite-Preview-04-17, and Claude 3.5 Sonnet. Gemini 2.0 Flash stands out within this group for having a comparable low token cost, while either leading or being among the best regarding the similarity measures. Moreover, its API provides high request limits, which we never exceeded in our tests. In contrast, we frequently hit rate limits when using DeepSeek and Mistral models. Based on performance and practical considerations, we recommend  Gemini 2.0 Flash. Among the openly available models, DeepSeek-V3 shows clearly the best results. 

Using Gemini 2.0 Flash, we translated all 19 thesauri with a total of 6475 terms, many with multilingual labels, in under 3.3 hours. This strongly suggests that WOKIE is well suited for small to medium-sized thesauri, as commonly created and used in the \gls{dh}. We also examined results for Latin separately, given its relevance in \gls{dh}. All LLM-enhanced combinations outperformed the best single primary service in all similarity measures. The best ones (GPT-4.1 mini, Gemini 2.0 Flash) achieved up to 0.31 string similarity, compared to 0.04 without LLM. This highlights the clear benefit of refinement for Latin.

\subsection{Ontology Matching}
WOKIE functions as a preprocessing step, therefore existing \gls{om} systems can benefit from translations without modifications. To answer the last research question regarding the impact of English pre-translation on \gls{om}, we used a multilingual SKOS benchmark from archaeology \cite{pourResultsOntologyAlignment2024}. All but one language were removed from the thesauri, then the matching was performed, and the alignment compared to ground truth alignments. We reproduced the benchmark using two existing matching systems and compared it to the results when pre-translation was performed with WOKIE. We used the F1 score (the harmonic mean of precision and recall) as our evaluation metric, as commonly done in \gls{om} benchmarking.

\begin{figure}
    \centering
    \includegraphics[width=0.85\linewidth]{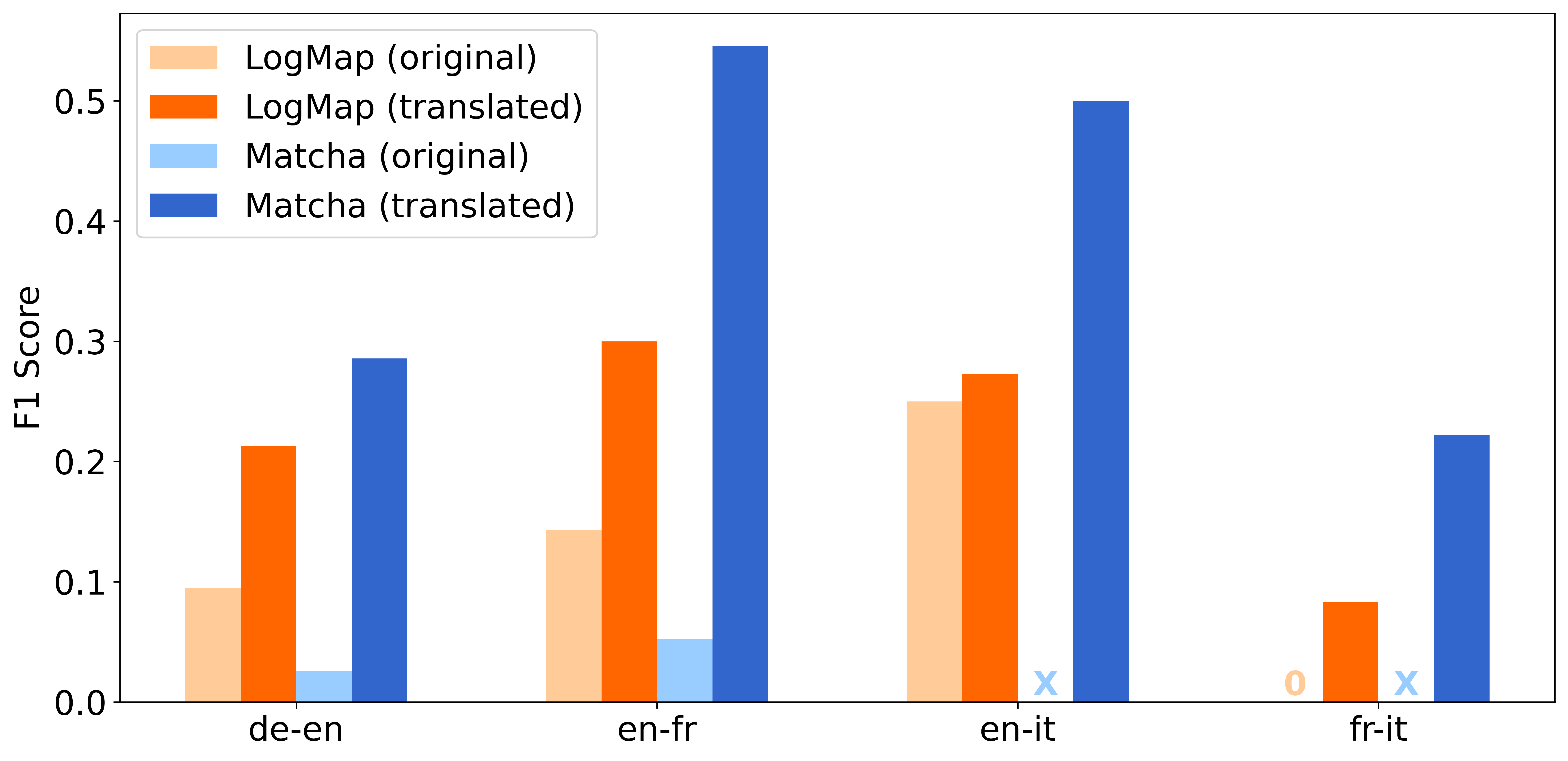}
    \caption{Comparison of the F1 scores from the original benchmark "(original)" with the results obtained after translating all thesauri to English first "(translated)". F1 scores of zero are denoted with a "0", errors on execution of the matching system with "X".}
    \label{fig:om}
\end{figure}

As shown in \cref{fig:om}, results improved for all language pairs after translation. The highest F1 score gain was observed using Matcha: an increase of 0.49 for English-French and a successful execution of previously failing English-Italian with an F1 score of 0.5. Even for the French-Italian pair, where both thesauri were translated, the F1 score for Matcha rises to 0.22. These findings demonstrate that introducing a preprocessing step for non-English thesauri can significantly improve matching results with minimal effort.

\section{Discussion and Limitations}
Despite our best efforts, testing all combinations of parameters, primary translation services, and \glspl{llm} was not feasible due to the large number of possible configurations and therefore exceeding rate limits. Additionally, it was not possible to include more languages, primarily due to the limited availability of high-quality multilingual thesauri. We showed that our pipeline is even applicable to the lower-resource language Serbian. Therefore, it is justified to assume that the pipeline also produces meaningful translations for untested languages, as long as they are supported by a subset of primary services and the selected LLM. Looking at domain specifics, we focused on those where domain-specific thesauri were available, and therefore do not claim to cover the wide range of domains belonging to  or using elements of the \gls{dh}. 

When examining multilingual labels used as ground truth, it becomes clear that this is not always justified. For example, in a thesaurus about early farming cultures, there is a term describing pottery decorations which has the English label "red on red (a3$\gamma$)" and a German label "Rot on red (a3c)". Our pipeline correctly translated it to "rot auf rot (a3$\gamma$)", which is nevertheless penalized because of an incorrect reference label. As a result, some systems may show lower similarity measures. It is also important to note that whenever automatic translation systems are involved, we recommend to either review the translations by experts or explicitly flag them as machine-generated. Looking at the quality of \gls{skos} thesauri in general, we located multiple issues when adhering to the data model. For instance, one thesaurus used URLs linking to Wikidata entities in the \texttt{skos:definition} field instead of the more appropriate \texttt{skos:related} or \texttt{skos:exactMatch} properties. Consequently, the pipeline performance is limited by the quality of the input data.

\section{Conclusions and Outlook}
In this work, we introduced WOKIE, a modular and ready-to-use translation pipeline designed to support the multilingual needs of the \gls{dh} community. WOKIE provides scalable translations of \gls{skos} thesauri without requiring prior knowledge of \gls{skos},  \gls{mt} or \glspl{llm}.  We systematically evaluated combinations of services, \glspl{llm} and parameter settings to identify the optimal configuration of WOKIE for \gls{dh} thesauri. 

Our evaluation demonstrated that WOKIE enables accurate, automatic and scalable translation of \gls{dh} thesauri. The integration of LLM-based refinement substantially improved translation quality, especially in ambiguous cases. We further showed that pre-translating thesauri into English prior matching significantly enhances the performance of ontology matching systems. In one benchmark, the F1-score rose from zero to 0.5 after translation. 

Looking at the scalability, we showed that translating thesauri with up to 10,000 terms is feasible when using services and models with sufficient request limits. To extend this further, we plan to integrate load balancer capabilities to handle even larger datasets.

WOKIE lowers technical barriers for multilingual vocabularies, supporting inclusive metadata practices and enabling better representation of communities with lower levels of English proficiency. Moreover, we plan to evaluate and possibly integrate synonyms provided by translations services. Since objective, automated, quantitative metrics for the translation quality of descriptions are difficult, if not impossible, we plan manual reviews including domain experts on small thesauri. 

Lastly, we aim to explore the impact on \gls{om} more broadly by investigating observed effects and testing additional matching systems with different datasets. To achieve this, we plan to include all suitable multilingual thesauri and matching systems participating in the recent campaigns of the \gls{oaei}. Overall, we believe that WOKIE contributes to more equitable and language-aware knowledge infrastructures, a key step toward inclusive and language-independent reuse of research data.

\section*{Acknowledgements}

This research was funded by the research program “Engineering Digital Futures” of the Helmholtz Association of German Research Centers, and the Helmholtz Metadata Collaboration Platform (HMC).

\paragraph{Disclosure.}Since the author's mother tongue is not English, \gls{llm} assisted tools such as LanguageTool
were used to improve punctuation, grammar, and spelling in full compliance with the Large Language Model Policy of the conference. In no way was any content created by these tools.

% Print the biblography at the end. Keep this line after the main text of your paper, and before an appendix. 
\printbibliography
\newpage

\printglossary[type=abbreviations]

\newpage

% You can include an appendix using the following command
\appendix

\section{Tables} \label{appdx:tables}

\begin{table}[!ht]
  \centering
  \begin{tabularx}{\linewidth}{@{} l c c Y @{}}
    \toprule
    Name 
      & Implemented? 
      & Free? 
      & Comment \\
    \midrule
        \href{https://translate.argosopentech.com}{Argos} & \ding{52} & \ding{52} & To be used locally with LibreTranslate (API of Argos) \\ \hline
        \href{https://cloud.google.com/translate/docs/reference/rest/}{Google Translate} & \ding{52} & \ding{54} & Uses Cloud Translation API \\ \hline
        \href{https://lingvanex.com}{Lingvanex} & \ding{52} & \ding{52} & ~ \\ \hline
        \href{{https://www.modernmt.com}ModernMT} & \ding{52} & \ding{52} & ~ \\ \hline
        \href{https://www.microsoft.com/en-us/translator}{Microsoft Translator} & \ding{52} & \ding{54} & Quite low request limits \\ \hline
        \href{https://de.pons.com/p/ubersetzungsapi/}{PONS} & \ding{52} & \ding{54} & No latin and serbian in contrast to the \href{https://de.pons.com/\%C3\%BCbersetzung}{PONS online dictionary} \\ \hline
        \href{https://www.reverso.net}{Reverso} & \ding{52} & \ding{52} & Sometimes no API response \\ \hline
        \href{https://yandex.com}{Yandex Translate} & \ding{52} & \ding{52} & ~ \\ \hline
        \href{https://www.alibabacloud.com/en/product/machine-translation?\_p\_lc=1}{Alibaba} & \ding{54} & ~ & Very slow \\ \hline
        \href{https://www.apertium.org}{Apertium} & \ding{54} & ~ & Unexpected exception when using the API \\ \hline
        \href{https://babelnet.org/}{BabelNet} & \ding{54} & ~ & Very low request limit \\ \hline
        \href{https://www.bing.com/translator}{Bing} & \ding{54} & ~ & Very low request limit \\ \hline
        \href{https://fanyi.caiyunapp.com/}{Caiyun} & \ding{54} & ~ & Very low requests per second \\ \hline
        \href{https://online.cloudtranslation.com}{CloudTranslation} & \ding{54} & ~ & Very low requests per second \\ \hline
        \href{https://developers.deepl.com/docs}{DeepL} & \ding{54} & ~ & Very low requests per second \\ \hline
        \href{https://elia.eus/start}{elia} & \ding{54} & ~ & Very slow \\ \hline
        \href{https://dict.hjenglish.com/app/trans}{hujiang} & \ding{54} & ~ & Very low requests per second \\ \hline
        \href{https://itranslate.com/webapp}{iTranslate} & \ding{54} & ~ & Very low requests per second \\ \hline
        \href{https://www.languagewire.com/en/technology/languagewire-translate}{languageWire} & \ding{54} & ~ & Very low requests per second \\ \hline
        \href{https://www.linguee.com}{Linguee} & \ding{54} & ~ & Very low requests per second \\ \hline
        \href{https://mymemory.translated.net}{Mymemory} & \ding{54} & ~ & Very low request limit \\ \hline
        \href{https://opennmt.net}{OpenNMT} & \ding{54} & ~ & Only for full texts \\ \hline
        \href{https://papago.naver.com}{Papago} & \ding{54} & ~ & Very slow, mainly for Korean \\ \hline
        \href{https://mt.qcri.org/api}{QcriTranslator} & \ding{54} & ~ & Obligatory registration failed \\ \hline
        \href{https://transmart.qq.com/}{qqTranSmart} & \ding{54} & ~ & Very slow \\ \hline
        \href{https://fanyi.sogou.com/text}{Sogou} & \ding{54} & ~ & Very slow \\ \hline
        \href{https://fanyi.qq.com}{Tencent} & \ding{54} & ~ & Identical to sogou \\ \hline
        \href{https://www.translate.com/machine-translation}{TranslateCom} & \ding{54} & ~ & Very low request limit \\ \hline
    \bottomrule
  \end{tabularx}
  \caption{Comparison of Translation APIs}
  \label{tab:translators}
\end{table}

Costs as they were on 1st of May 2025 using the API provided by the manufacturer.

\begin{table}[!ht]
  \centering
  \small
  \begin{tabularx}{\linewidth}{@{} p{3cm} p{1cm} p{1.5cm} p{1.5cm} p{1cm} X @{}}
    \toprule
        Model Name &
        Imple- mented? &
        Input Costs (USD/1M tk) & 
        Costs for about 100 terms in ¢\footnotemark{} & 
        Free to download? & 
        Comment \\
    \midrule
        claude-3-5-haiku & \ding{52} & 0.80 & 2.4 &  \ding{54} & ~ \\ \hline
        claude-3-5-sonnet & \ding{52} & 3  & 9 &  \ding{54} & ~ \\ \hline
        claude-3-7-sonnet & \ding{52} & 3  & 9 &  \ding{54} & ~ \\ \hline
        claude-3-haiku & \ding{52} & 0.25  & 0.75 &  \ding{54} & ~ \\ \hline
        codestral-latest & \ding{52} & 0.3  & 0.9 &  \ding{54} & Pointing to Codestral-2501 \\ \hline
        deepseek-chat & \ding{52} & 0.27  & 0.81 &  \ding{52} & Pointing to DeepSeek-V3 \\ \hline
        deepseek-reasoner & \ding{52} & 0.14  & 0.42 &  \ding{52} & Pointing to DeepSeek-R1 \\ \hline
        gemini-1.5-flash & \ding{52} & 0.075  & 0.225 &  \ding{54} & ~ \\ \hline
        gemini-1.5-flash-8b & \ding{52} & 0.0375  & 0.1125 &  \ding{54} & ~ \\ \hline
        gemini-2.0-flash & \ding{52} & 0.10  & 0.3 &  \ding{54} & ~ \\ \hline
        gemini-2.0-flash-lite & \ding{52} & 0.075  & 0.225 &  \ding{54} & ~ \\ \hline
        gemini-2.5-flash-preview-04-17 & \ding{52} & 0.15  & 0.45 &  \ding{54} & ~ \\ \hline
        gemma3:12b & \ding{52} & 0 & 0  &  \ding{52} & ~ \\ \hline
        gpt-3.5-turbo & \ding{52} & 0.50  & 1.5 &  \ding{54} & ~ \\ \hline
        gpt-4.1-mini & \ding{52} & 0.40  & 1.60 &  \ding{54} & ~ \\ \hline
        gpt-4.1-nano & \ding{52} & 0.10  & 0.3 &  \ding{54} & ~ \\ \hline
        gpt-4.1 & \ding{52} & Unknown  & Unknown &  \ding{54} & ~ \\ \hline
        gpt-4o & \ding{52} & 2.50 & 7.5 &  \ding{54} & ~ \\ \hline
        gpt-4o-mini & \ding{52} & 0.15  & 0.45 &  \ding{54} & ~ \\ \hline
        llama-4-maverick:free & \ding{52} & 0 & 0 &  \ding{52} & ~ \\ \hline
        ministral-3b-latest & \ding{52}  & 0.04 & 0.12 &  \ding{54} & Pointing to Ministral 3B-2410 \\ \hline
        mistral-large-latest & \ding{52} & 2  & 6 &  \ding{54} & Pointing to Mistral Large-2411 \\ \hline
        mistral-medium-latest & \ding{52} & 0.4  & 1.2 &  \ding{54} & Pointing to Mistral Medium-2312 \\ \hline
        mistral-tiny-latest & \ding{52} & Unknown &  ~ & \ding{54} & Pointing to Mistral Tiny-2407 \\ \hline
        mistral-small-latest & \ding{52} & 0.15  & 0.45 &  \ding{52} & Pointing to Mistral Small-2503 \\ \hline
        open-mistral-nemo & \ding{52} & 0.15 & 0.45 &  \ding{52} & ~ \\ \hline
        open-mixtral-8x22b & \ding{52} & 2  & 6 &  \ding{52} & ~ \\ \hline
        codestral-mamba-latest & \ding{54} & Unknown & Unknown &  \ding{52} & Poor performance; Pointing to Codestral Mamba \\ \hline
        llama-4-scout:free & \ding{54} & 0 & 0 &  \ding{52} & Cannot be reliably constrained to adhere to the expected output format \\ \hline
        llama3.2 (ollama) & \ding{54} & 0 & 0 &  \ding{52} & Returned output is irrelevant and lacks meaningful content \\ \hline
        ministral-8b-latest & \ding{54} & 0.1 & 0.3 &  \ding{54} & Poor performance; Pointing to Ministral 8B-2410 \\ \hline
        open-codestral-mamba & \ding{54} &  ~ & ~ & \ding{52} & Cannot be reliably constrained to adhere to the expected output format, also overly verbose \\ \hline
        open-mistral-7b & \ding{54} &  ~ &  ~& \ding{52} & Cannot be reliably constrained to adhere to the expected output format, also overly verbose \\ \hline
        open-mixtral-8x7b & \ding{54} & 0.7  & 0.21 &  \ding{52} & Cannot be reliably constrained to adhere to the expected output format, also overly verbose \\ \hline
    \bottomrule
  \end{tabularx}
  \caption{Comparison of \glspl{llm}}
  \label{tab:llms}
\end{table}
\addtocounter{footnote}{0}
\footnotetext{The input tokens dominate the costs largely in WOKIE, which is why the total costs for a thesaurus are only calculated with the input costs.}

\begin{table}[!ht]
  \centering
  \small
  %\begin{tabularx}{\linewidth}{@{} p{3cm} p{3cm} p{3cm} @{}}
  \begin{tabularx}{\linewidth}{@{} X X X @{}}
    \toprule
        Original label (en) & Original label (de) & Translated label by WOKIE (de) \\ 
        \midrule
        Analyzing & Analyse & Analyse \\ \hline
        Annotating & Annotieren & Anmerkung \\ \hline
        Archiving & Archivieren & Archivierung \\ \hline
        Capturing & Erfassen & Erfassung \\ \hline
        Collaborating & Kollaboration & Zusammenarbeit \\ \hline
        Commenting & Kommentieren & Kommentar \\ \hline
        Communicating & Kommunizieren & Kommunikation \\ \hline
        Content Analysis & Inhaltsanalyse & Inhaltsanalyse \\ \hline
        Contextualizing & Kontextualisieren & Kontextualisierung \\ \hline
        Converting & Konvertieren & Konvertierung \\ \hline
        Creating & Erzeugen & Schöpfung \\ \hline
        Crowdsourcing & Crowdsourcing & Crowdsourcing \\ \hline
        Data Cleansing & Bereinigen & Datenbereinigung \\ \hline
        Data Recognition & Datenerkennung & Datenerkennung \\ \hline
        Designing & Design & Design \\ \hline
        Discovering & Entdecken & Entdeckung \\ \hline
        Disseminating & Dissemination & Verbreitung \\ \hline
        Editing & Bearbeiten & Bearbeitung \\ \hline
        Enriching & Anreichern & Anreicherung \\ \hline
        Gathering & Sammeln & Sammlung \\ \hline
        Identifying & Identifizieren & Identifizierung \\ \hline
        Imaging & Bilderfassung & Bildgebung \\ \hline
        Interpreting & Interpretation & Interpretation \\ \hline
        Modeling & Modellieren & Modellierung \\ \hline
        Network Analysis & Netzwerkanalyse & Netzwerkanalyse \\ \hline
        Organizing & Organisieren & Organisieren \\ \hline
        Preserving & Konservierung & Erhaltung \\ \hline
        Programming & Programmieren & Programmierung \\ \hline
        Publishing & Veröffentlichen & Veröffentlichung \\ \hline
        Recording & Aufzeichnen & Aufnahme \\ \hline
        Relational Analysis & Analyse von Relationen & Relationale Analyse \\ \hline
        Sharing & Teilen & Teilen \\ \hline
        Spatial Analysis & Räumliche Analyse & Raumanalyse \\ \hline
        Storing & Speicherung & Speicherung \\ \hline
        Structural Analysis & Strukturanalyse & Strukturanalyse \\ \hline
        Stylistic Analysis & Stilistische Analyse & Stilanalyse \\ \hline
        Theorizing & Theoriebildung & Theoretisierung \\ \hline
        Transcribing & Transkription & Transkription \\ \hline
        Translating & Übersetzen & Übersetzung \\ \hline
        Visual Analysis & Visualisierung & Visualisierung \\ \hline
        Web Development & Webentwicklung & Webentwicklung \\ \hline
        Writing & Schreiben & Schreiben \\ \hline
    \bottomrule
  \end{tabularx}
  \caption{Original English, original German and translated German labels of TaDiRAH. The translation was obtained by WOKIE without knowledge of the original German label. }
  \label{tab:exemplary-translations}
\end{table}

\end{document}